# Predicting the Outcome of Judicial Decisions made by the European Court of Human Rights

Conor O'Sullivan, Joeran Beel

School of Computer Science and Statistics, Trinity College, Ireland
ADAPT Centre
osullc43@tcd.ie, beelj@tcd.ie

**Abstract.** In this study, machine learning models were constructed to predict whether judgements made by the European Court of Human Rights (ECHR) would lead to a violation of an Article in the Convention on Human Rights. The problem is framed as a binary classification task where a judgement can lead to a "violation" or "non-violation" of a particular Article. Using auto-sklearn, an automated algorithm selection package, models were constructed for 12 Articles in the Convention. To train these models, textual features were obtained from the ECHR Judgment documents using N-grams, word embeddings and paragraph embeddings. Additional documents, from the ECHR, were incorporated into the models through the creation of a word embedding (echr2vec) and a doc2vec model. The features obtained using the echr2vec embedding provided the highest cross-validation accuracy for 5 of the Articles. The overall test accuracy, across the 12 Articles, was 68.83%. As far as we could tell, this is the first estimate of the accuracy of such machine learning models using a realistic test set. This provides an important benchmark for future work. As a baseline, a simple heuristic of always predicting the most common outcome in the past was used. The heuristic achieved an overall test accuracy of 86.68% which is 29.7% higher than the models. Again, this was seemingly the first study that included such a heuristic with which to compare model results. The higher accuracy achieved by the heuristic highlights the importance of including such a baseline.

## 1 Introduction

The European Court of Human Rights (ECHR) is an international court that examines potential breaches of the European Convention on Human Rights. The Convention consists of numerous Articles. For example, Article 2: Right to life and Article 6: Right to a fair trial [5]. According to [6], for a potential breach of the Convention to be investigated an application must first be made. This means the ECHR cannot investigate potential violations on its own accord. Any State or individual can make an application but cases can only be made against one of the 47 States that has ratified the Convention. Since its founding, the Court has been very successful leading to a growing number of cases. In the Court's own words:

"The Court has been a victim of its own success: over 50,000 new applications are lodged every year. The repercussions of certain judgments of the Court, on a regular basis, and the growing recognition of its work among nationals of the States Parties, have had a considerable impact on the number of cases brought every year [6, p. 7]."

The problem is that the large number of applications made every year has led to a backlog of applications. This has subsequently led to significant time delays in Court proceedings. Due to this backlog, applications can take up to a year before an initial examination can take place. After this examination, the application has to go through a further process before the Court can determine whether there was a breach of the Convention [7]. Ultimately, it can take over a year for the ECHR to make a final judgement.

This paper seeks to determine how accurately can the judgements made by the ECHR be predicted. This is done using final Judgment documents, produced by the ECHR, as input [1]. Using Natural Language Processing (NLP) techniques, textual features are obtained from these documents. Machine learning models are then trained, using these features, to predict whether an application has resulted in a "violation" or "non-violation" of a human right. Ultimately, a predictive model can be used to help address the backlog of applications. The ECHR could use an accurate predictive model to make or help make judgements. Such a model could also be used to prioritise cases. That is cases which indicate a high likelihood of violation can be prioritised.

## 2 Related Work

Table 1 shows the results of studies that looked at predicting the outcome of legal cases. The "Court" column gives the legal court considered by the study. The majority of the studies looked at either the ECHR or the Supreme Court of the United States (SCOTUS). The "Data" column in Table 1 gives the type of data used in the study. "Case documents" refers to text documents that outline the cases heard by the Court. The majority of the studies that looked at SCOTUS decisions used "Summary Information" [22] [10] [11] [12]. These are variables that summarise the cases. Additionally, in Table 1 the "Target Variable" is usually a simplification of potential case outcomes. For each study, the "Algorithm" gives the algorithm that achieved the highest accuracy when predicting the target variable. The "Train. Acc." and "Test Acc." give the training and test accuracy, respectfully, achieved by the study. 'For all the studies, the training accuracy is the cross-validation accuracy achieved by the study.

For the work done on the ECHR, there are some potential issues with the datasets used. For instance, in their final training sets, [3] had 250, 80 and 254 cases for Articles 3, 6 and 8 respectfully. The number of Judgments for Article 6 is peculiar as according to [6] the majority of the Judgments made by the ECHR involved Article 6. This is not the case for the Judgments collected by [3] and this suggests their dataset is not representative of all Judgments. The issues with dataset collection could be a result of how data has been made available by the ECHR. It is not possible to download Judgments and other documents from the HUDOC database in bulk [1]. This means researchers have to create their own tools to download the data which can be unreliable. For example, [18, p .8] states: "We used a rather crude automatic extraction method, so it is possible that a few cases might be missing from our dataset."

[1] An example of a Judgment: `http://hudoc.echr.coe.int/eng?i=001-194614`

**Table 1.** Summary of Previous Works

| Author | Court | Data | Target Variable | Algorithm | Train. Acc. | Test Acc. |
|---|---|---|---|---|---|---|
| [3] | ECHR | Case Documents | Violation, Non-Violation | SVM | 80.1% | NA |
| [15] | ECHR | Case Documents | Violation, Non-Violation | SVM | 79.5% | NA |
| [18] | ECHR | Case Documents | Violation, Non-Violation | SVM | 75.0% | 74.0% |
| [17] | SCOTUS | Summary Information | Affirmed, Reversed | Decision Tree | NA | 75% |
| [10] | SCOTUS | Sumamry Information: Justice Votes | Justice Decsion: Affirmed, Reversed | Stochastic Block Model | NA | 83% |
| [11] | SCOTUS | Summary Information | Affirmed, Reversed | Random Forest | NA | 70% |
| [12] | SCOTUS | Summary Information and Oral Arguments | Affirmed, Reversed | Random Forest | 74.04% | NA |
| [2] | US Circuit Court | Case Documents | Affirmed, Reversed | CNN | 79% | NA |
| [2] | SCOTUS | Case Documents | Affirmed, Reversed | Random Forest | 68% | NA |

An important aspect of the machine learning process is to include a test set. Models can become biased towards the training set or, in other words, they have been over-fitted to the training set. So by including a test set, we obtain an unbiased estimate of how well the models perform [13, p. 67]. Additionally, to obtain a realistic estimate of the model's performance a realistic test set should be used. Where a realistic set is one where the target variables are in the same proportion to what we would expect in the future. For example, [22] trained their model using SCOTUS cases before the 2002 term. This was done before the start of the 2002 term. Once the term started, the researchers tested their models on the cases, as they transpired, throughout the term. This is inherently a realistic test set as the proportion of "affirmed" and "reversed" cases are the same as in reality. For the ECHR, a less elaborate way of obtaining a realistic test set would be to choose the set so that it had the same proportion of "violation" to "non-violations" as in the past. This is assuming that future Judgments will have a similar proportion.

In Table 1, we see that neither [3] nor [15] have included a test set. [18] has included a test set but it is not realistic. [18] has used a balanced training set for each Article. They obtained the largest training sets possible. For example, Article 6 has more violations than non-violations and so the training set, for this Article, contains all the non-violations. The remaining Judgments are used as the test set. Ultimately, what this means is that, depending on the Article, the test sets contain only either violation or non-violations and not both. Consequentially, as far as we can tell, no study that looks at predicting ECHR judgements has used a realistic test set to evaluate their models. In terms of the research question, this means that we do not have a realistic estimate of how well machine learning models can predict the judgements made by the ECHR.

[3], [15] and [18] have used N-gram features to train their models. Specifically, they have obtained these textual features from ECHR Judgment documents. These features have their limitations. For instance, they do not consider the semantics of words. The word order of the legal documents is also lost [14]. As an alternative, word embeddings could be used to obtain features. Features could be obtained using the pre-trained legal embedding, law2vec. These embeddings have been trained on a variety of legal documents [4]. As a result, these embeddings could have captured the legal semantics of words. Another option would be to train a new word embedding using additional documents obtained from the ECHR. This is one way of incorporating additional ECHR data, not just Judgments, into the machine learning models. Additionally, paragraph embeddings could be used as an alternative. These are similar to word embeddings but they consider the word order of documents [14].

## 3 Methods

### 3.1 Data Preparation

All ECHR documents were obtained using an API provided by vizlegal. vizlegal is a legal technology company that specialises in legal search [23]. Ultimately, the vizlegal API was used as it was the most reliable and efficient method of obtaining the data that could be found. The number of documents, downloaded using the API, are shown in Figure 1. Decisions make up approximately 40% of all the documents. These documents give the rulings on the admissibility of applications. Communicated cases describe the communications that took place with the State responding to an application. Legal summaries are summaries of important judgements or decisions and the resolution documents describe proposals made by the Court [8]. The other documents include Reports and Advisory Opinions. In total, 56688 documents were obtained. Out of these, 14071 Judgment documents were obtained. As this study attempts to predict the outcome the outcome of judgements, the Judgment documents are the primary data source. The other documents are still incorporated through the process of creating word and paragraph embeddings.

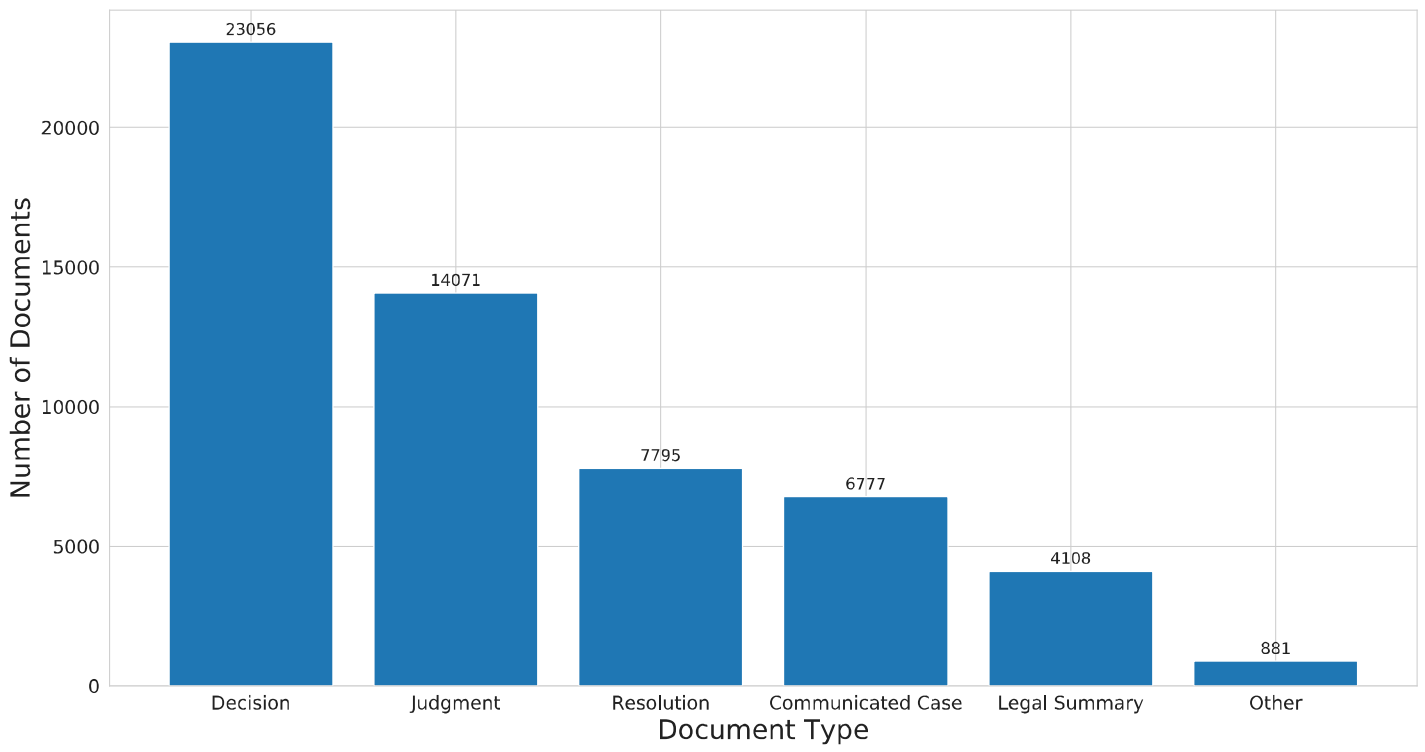

**Fig. 1.** Number of ECHR Documents

Only Judgments with a specific structure are considered. That is they must have a procedure, facts, law and verdict section. The facts section must also consist of two subsections: circumstances and relevant law. 9703 of the 14071 Judgment have this structure. The choice was made as all previous papers discussed have decided to use only Judgments of this structure [3] [15] [18]. This is because a standard document structure simplifies the process of cleaning and extracting different textual features from the documents.

The 9703 Judgments are put into groups based on what Articles they address. The Judgments are then labelled as "non-violation" if the there were no violations for that Article and "violation" if there was atleast 1 violation for the Article. Ultimately, the problem has been framed as a binary classification problem with respect to each Article. A balance training set is then selected. This is to avoid models becoming biased towards one of the outcomes. Realistic test sets were selected. That is, they were chosen so that they had the same violation to non-violation ratio as seen in the past. The final number of Judgments in the training and test set for each article can be seen in Table 2.

### 3.2 Feature Engineering

From each Judgment, the text from the procedure, facts, circumstances and relevant sections are obtained. A combination of the text from the procedure and facts sections (procedure+facts) is also created. To avoid data leakage, the text from the law and verdict section is not used as they contain details of the case verdicts. The text from each section is then cleaned by making it lower case and removing all punctuation and numbers. A version of this text with stop-words removed is also obtained. The set of English stop-words provided by the NLTK package was used [16]. Each corpus is divided into a training and test set. Different textual features are subsequently created using these different corpora.

**Table 2.** Number of Judgments in Training and Testing Sets

| | Training Set | | Testing Set | |
|---|---|---|---|---|
| **Article** | **Violation** | **Non-violation** | **Violation** | **Non-violation** |
| **Article 2** | 76 | 76 | 58 | 8 |
| **Article 3** | 245 | 245 | 175 | 27 |
| **Article 5** | 166 | 166 | 168 | 18 |
| **Article 6** | 504 | 504 | 539 | 56 |
| **Article 7** | 32 | 32 | 4 | 5 |
| **Article 8** | 271 | 271 | 93 | 30 |
| **Article 9** | 20 | 20 | 5 | 2 |
| **Article 10** | 128 | 128 | 45 | 14 |
| **Article 11** | 23 | 23 | 14 | 3 |
| **Article 13** | 101 | 101 | 138 | 11 |
| **Article 14** | 182 | 182 | 20 | 20 |
| **Article 18** | 13 | 13 | 1 | 2 |

**N-grams** The NLTK package is used to create N-gram feature matrices using the cleaned Judgment text [16]. From the Judgments in the training set, the 2000 most frequent N-grams of length 1 to 4 are obtained. Using the training set N-grams, the Judgments are vectorised to obtain feature matrices for both the training and test sets. An example of the resulting matrix can be seen in Figure 2. Here, the rows represent the individual Judgments and the columns give the frequency of the particular N-gram for that Judgment. These feature matrices are then normalised using Min-Max feature scaling.

| | abduction | ability | able | abroad | ... | years imprisonment | years old | yet | zone | target |
|---|---|---|---|---|---|---|---|---|---|---|
| **0** | 0 | 0 | 0 | 0 | ... | 0 | 0 | 0 | 0 | 0 |
| **1** | 2 | 0 | 3 | 0 | ... | 1 | 0 | 3 | 0 | 0 |
| **2** | 0 | 1 | 2 | 1 | ... | 0 | 0 | 1 | 0 | 0 |
| **3** | 0 | 0 | 0 | 0 | ... | 0 | 0 | 0 | 0 | 0 |
| ⋮ | ⋮ | ⋮ | ⋮ | ⋮ | ⋮ | ⋮ | ⋮ | ⋮ | ⋮ | ⋮ |
| **538** | 0 | 0 | 0 | 0 | ... | 0 | 0 | 0 | 0 | 1 |
| **539** | 1 | 0 | 0 | 0 | ... | 2 | 0 | 1 | 0 | 1 |
| **540** | 0 | 7 | 0 | 0 | ... | 3 | 0 | 2 | 0 | 1 |
| **541** | 0 | 0 | 0 | 0 | ... | 0 | 0 | 0 | 0 | 1 |

**Fig. 2.** N-gram Feature Matrix Example

**Word Embeddings** The different word embeddings used are summarised in Table 3. 'Corpus' gives the documents that are used to train the word embeddings and 'No. Tokens' are the number of lower case words that make up the corpus. The 'vocabulary size' is the number of words that have vector representations for that embedding. For each embedding, a 100 dimension and 200 dimension version are used. The GloVe embeddings were trained by [20] and the law2vec embeddings were trained by [4]. The echr2vec embeddings were trained specifically for this paper.

**Table 3.** Summary of Word Embeddings

| Embedding | Corpus | No. Tokens | Vocabulary Size |
|---|---|---|---|
| GloVe | Gigaword 5 and Wikipedia 2014 | 6 Billion | 400,000 |
| law2vec | 123,066 legislation documents | 492 Million | 169,439 |
| echr2vec | ECHR Judgment documents | 84 Million | 47,587 |

When creating the echr2vec embeddings, all of the 56688 ECHR documents obtained were considered. However, to avoid data leakage it was necessary to exclude certain documents or sections of documents. Ultimately, this means that the echr2vec embedding was not trained on any text that would not be available before a judgement was made. The embedding was created using these documents and the gensim implementation of the word2vec model [21]. A 5-word window and a minimum threshold of 10 occurrences were used.

**Average Embedding Values** Word embeddings allow us to represent words as vectors. For this problem, predictions are made using entire documents so it is necessary to represent the documents as vectors. This is done by finding the average of the word embedding vectors. From this process, a feature matrix similar to the one in Figure 2 is obtained. As before, the rows represent each Judgment. Except now, the columns represent each element of the average word vector.

**Doc2vec Embeddings** A doc2vec model has also been used to represent the Judgments as vectors. The gensim implementation was used to train the doc2vec model [21]. The same documents used to train the echr2vec embeddings were used. The models were trained using a window size of 15, a minimum word frequency of 10 and 20 training epochs. Judgment vectors are inferred using 20 epochs and both 100 and 200 dimension vectors are obtained.

### 3.3 Modelling

Using the textual features discussed above, machine learning models are constructed for each of the Articles. This was done using the auto-sklearn python package [9]. It is based on the scikit-learn machine learning framework [19] and it considers 15 different algorithms, including Linear SVM, Gradient Boosting and Random Forest. 10-fold cross-validation is used to select both the algorithm and the associated hyper-parameters. Ultimately, this package provides an alternative to grid search and a wider range of models and parameters can be tested than tested in previous papers [9].

A distinction should be made between the hyper-parameters in Table 4 and those selected by the auto-sklearn package. The auto-sklearn package can only be used to select the algorithm and the algorithm's associated hyper-parameters. Hence, for each of an Article's feature matrices, auto-sklearn is used to find the classification algorithm and associated hyper-parameters that maximises cross-validation accuracy. Then, all these cross-validation accuracies are compared to obtain the model with the highest overall cross-validation accuracy.

**Table 4.** Model Hyper-parameters

| Hyper-parameter | Values |
|---|---|
| Feature Type | N-gram, GloVe,,law2vec, echr2vec, doc2vec |
| Dimension | 100, 200 and 2000 (for N-gram only) |
| Judgment Section | procedure+facts ,procedure, facts, circumstances, relevant |
| Stop-words | Yes, No |

Ultimately, at the end of this process, we will have one model for each Article. This model would have been trained using one combination of the hyper-parameters in 4. The classification algorithm and it's associated algorithm would have been selected by the auto-sklearn package. For each Article, this model is then re-trained using the entire training set and used to make predictions on the test set. This is to provide an estimation of how well the model performs on a realistic out-of-sample data set. The models' results are also compared to a simple heuristic. That is, the heuristic always predicts the outcome of the Judgment to be the outcome that was the most common in the past. For example, Article 6 has had more violations than non-violations. Each Judgment in the test set for Article 6 will, consequentially, be predicted as a violation by the heuristic.

## 4 Results

The models achieved a weighted average of 0.6883 across all the Articles. Where the weights are given by the number of Judgments in the test set for each Article. This is the best estimation of how well the models will perform in a realistic scenario on new cases. The accuracy of the models and the heuristic on the test set can be seen in Figure 3. The accuracy for each Article as well as the weighted average across all the Articles are shown.

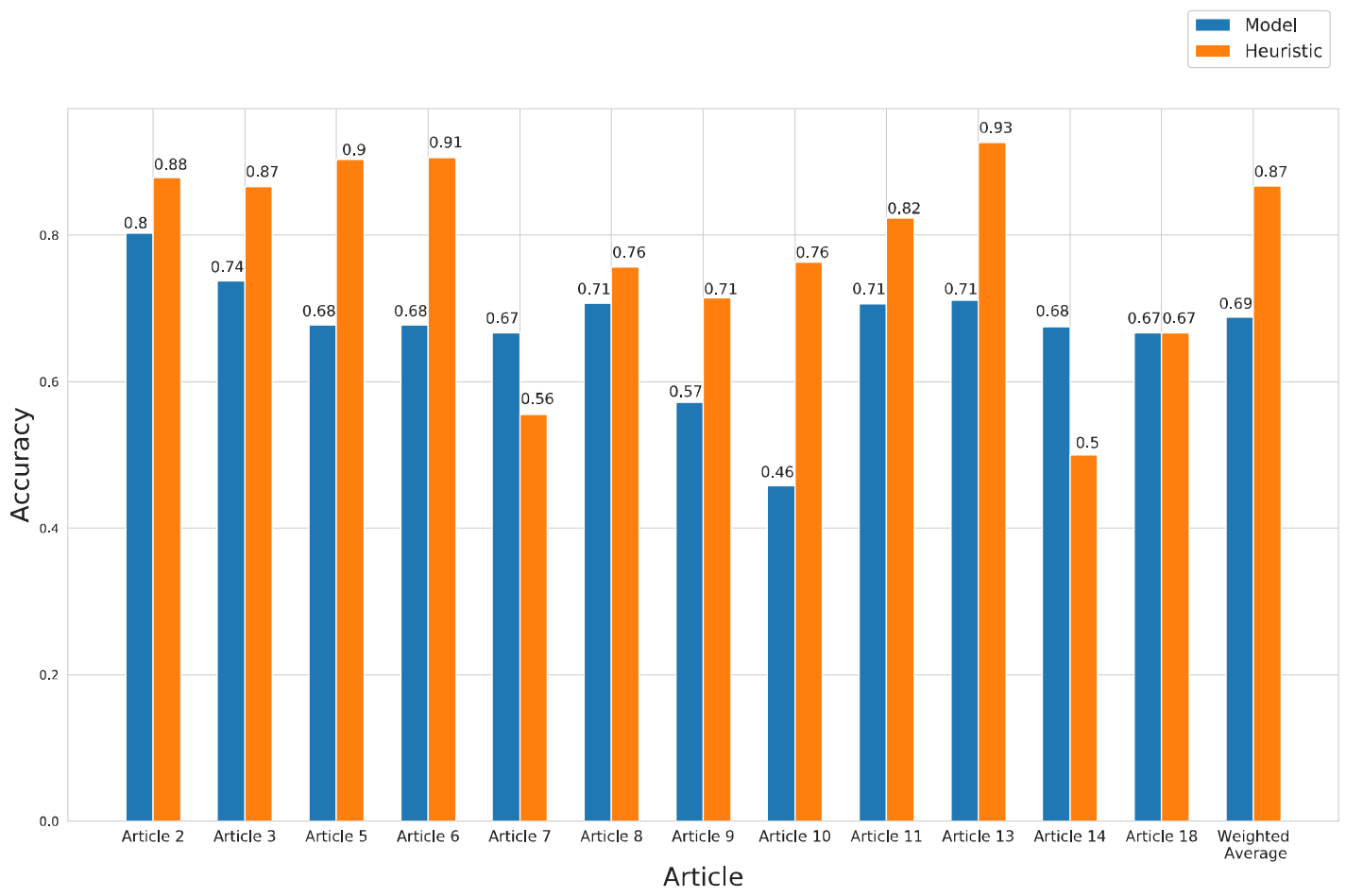

**Fig. 3.** Model and Heuristic Accuracy on Test Set

The test accuracies in Figure 3 can be compared to the heuristic accuracies. For all Articles, excepting 7, 14 and 18, the accuracy of the heuristic on the test set was higher. The weighted average, for the heuristic, was 0.8668 which is 29.7% higher than the weighted average for the models on the test set. Hence, in general, the heuristic has outperformed the models.

The hyper-parameters and classification algorithm that achieved the highest cross-validation accuracy for each Article can be seen in Table 5. The "Feature Type", "Dimension", "Section" and "Stopwords" parameters are discussed in the Methods section. The "Classifier" is the classification algorithm that was selected by the auto-sklearn package. In the Related Work section, we saw that a linear SVM produced the highest cross-validation accuracy for those papers that looked at the ECHR. Looking at Table 5, we can see that, in this study, a linear SVM did not produce the highest cross-validation accuracy for any of the Articles. This is important as it suggests that, to improve accuracy, it was necessary to test additional classification algorithms.

**Table 5.** Model Hyper-parameters

| **Article** | **Feature Type** | **Dimension** | **Section** | **Stopwords** | **Classifier** |
|---|---|---|---|---|---|
| Article 2 | law2vec | 100 | circumstances | Yes | Gradient Boosting |
| Article 3 | GloVe | 200 | procedure+facts | No | Random Forest |
| Article 5 | GloVe | 200 | relevant | Yes | Gradient Boosting |
| Article 6 | echr2vec | 100 | procedure+facts | Yes | SGD |
| Article 7 | GloVe | 200 | circumstances | No | Decision Tree |
| Article 8 | echr2vec | 100 | procedure+facts | Yes | Random Forest |
| Article 9 | n-gram | 2000 | circumstances | Yes | AdaBoost |
| Article 10 | echr2vec | 200 | procedure+facts | No | QDA |
| Article 11 | GloVe | 200 | procedure+facts | Yes | SGD |
| Article 13 | GloVe | 100 | procedure+facts | Yes | QDA |
| Article 14 | echr2vec | 200 | procedure+facts | No | QDA |
| Article 18 | echr2vec | 200 | procedure+facts | No | Random Forest |

The higher accuracy of the heuristic can be partly explained by the balance of violation to non-violations in the test sets. Take for instance Article 6 where, in the past, 91% of the complaints about this Article resulted in violations. As the test sets had the same balance as past judgements, the heuristic correctly predicted the outcome of Article 6 judgements with 91%. The recall and precision of the models further explain why the heuristic outperformed the models. Figure 4, shows the precision and recall, on the test sets, of the models. 7 of the Articles had a precision above 0.9 and 9 of the Articles had a precision above 0.8. In general, the high precisions mean that models tend not to miss-classify non-violations as violations. In comparison, lower recall values are observed. For 9 of the Articles, the precision was higher than the recall and the average recall was 0.6906. The lower recall, means the models tend to miss-classify violation cases as non-violation cases. In other words, incorrect predictions are mainly due to false negatives.

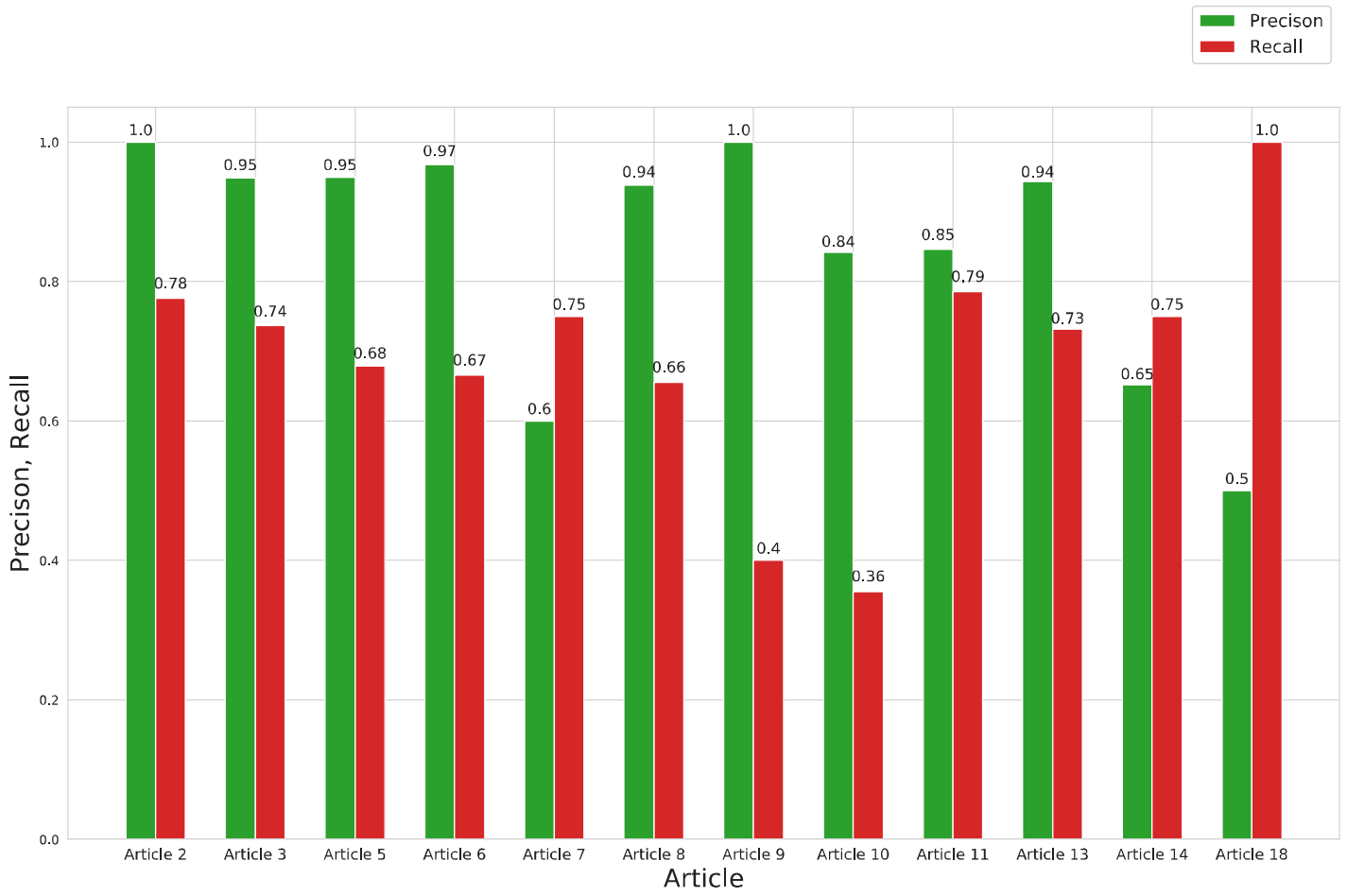

**Fig. 4.** Accuracy, Precision and Recall on Test Set

The models could still be used to prioritise cases by identifying which cases are more likely to lead to violations. The heuristic does not provide any benefit in terms of prioritising cases. As the predictions for each Article would be the same, all complaints would be given the same priority. In this sense, the models may be more useful. As discussed above, the tendency to have a high precision means there are relatively few false positives. This means the cases identified as violations and subsequently prioritised, will tend to be violations. The downside is that those judgements, misclassified as non-violations, would be given equal priority to the remaining non-violation cases. Nonetheless, overall the models would put the cases in a better order as more violation cases would be heard sooner.

## 5 Conclusion

Given the results of the models, it is unlikely that the ECHR would use the models to make judgements. Using a realistic data set, the models achieved a weighted average of 68.83% across all the Articles. Where the weights are given by the number of Judgments in the test set for each Article. Hence, it is estimated that if the models are used by the ECHR over 30% of rulings on human rights violations would be incorrect. The consequences of this could be severe considering that the Court was set up to protect human rights. As discussed, the models could still be a useful tool. The models could provide an indication of which applications in the backlog should be prioritised.

Ultimately, the research conducted is not enough to solve the research problem. Nonetheless, the study has made some contributions to this area of research. As far as we could tell, the first realistic test set has been used to determine the accuracy of the models. This provided the first realistic estimate of how well machine learning algorithms can predict the outcome of judgements made by the ECHR. This is an important baseline that the results of future work can be compared to.

A limitation of this study is that the models constructed provided only the final predictions for each Judgment. They did not provide any indication of how predictions are made. In reality, Judges have to justify their decisions and so they would not be able to rely on a model that gives only a final prediction. In addition to improving accuracy, this is an aspect of the models that should be considered. Models that provide information on how predictions are made would likely be more useful to Judges.

**Acknowledgement.** This research was partially conducted at the ADAPT SFI Research Centre at Trinity College Dublin. The ADAPT SFI Centre for Digital Media Technology is funded by Science Foundation Ireland through the SFI Research Centres Programme and is co-funded under the European Regional Development Fund (ERDF) through Grant # 13/RC/2106.

## References

1. Hudoc database. https://echr.coe.int/Pages/home.aspx?p=caselaw/HUDOC&c= (2018), accessed: 2018-11-18
2. Agrawal, S., Ash, E., Chen, D., Gill, S.S., Singh, A., Venkatesan, K.: Affirm or reverse? using machine learning to help judges write opinions. Tech. rep., Working Paper (2017)
3. Aletras, N., Tsarapatsanis, D., Preoţiuc-Pietro, D., Lampos, V.: Predicting judicial decisions of the european court of human rights: A natural language processing perspective. PeerJ Computer Science 2, e93 (2016)
4. Chalkidis, I., Kampas, D.: Deep learning in law: early adaptation and legal word embeddings trained on large corpora. Artificial Intelligence and Law 27(2), 171–198 (2019)
5. Council of Europe: European convention for the protection of human rights and fundamental freedoms, as amended by protocols nos. 11 and 14. https://www.refworld.org/docid/3ae6b3b04.html (1950), accessed: 2019-06-30
6. Council of Europe: European court of human rights: The echr in 50 questions. https://www.echr.coe.int/Documents/50Questions_ENG.pdf (2014), accessed: 2019-06-30

7. Council of Europe: European court of human rights: Questions and answers. https://www.echr.coe.int/Documents/Questions_Answers_ENG.pdf (2016), accessed: 2019-06-30
8. Council of Europe: European court of human rights: Hudoc user manual. https://www.echr.coe.int/Documents/HUDOC_Manual_ENG.PDF (2017), accessed: 2019-06-30
9. Feurer, M., Klein, A., Eggensperger, K., Springenberg, J., Blum, M., Hutter, F.: Efficient and robust automated machine learning. In: Cortes, C., Lawrence, N.D., Lee, D.D., Sugiyama, M., Garnett, R. (eds.) Advances in Neural Information Processing Systems 28, pp. 2962–2970. Curran Associates, Inc. (2015), http://papers.nips.cc/paper/5872-efficient-and-robust-automated-machine-learning.pdf
10. Guimerà, R., Sales-Pardo, M.: Justice blocks and predictability of us supreme court votes. PloS one 6(11), e27188 (2011)
11. Katz, D.M., Bommarito II, M.J., Blackman, J.: A general approach for predicting the behavior of the supreme court of the united states. PloS one 12(4), e0174698 (2017)
12. Kaufman, A., Kraft, P., Sen, M.: Machine learning, text data, and supreme court forecasting. Project Report, Harvard University (2017)
13. Kuhn, M., Johnson, K.: Applied Predictive Modeling. Springer, Springer New York Heidelberg Dordrecht London (2013)
14. Le, Q., Mikolov, T.: Distributed representations of sentences and documents. In: International conference on machine learning. pp. 1188–1196 (2014)
15. Liu, Z., Chen, H.: A predictive performance comparison of machine learning models for judicial cases. In: Computational Intelligence (SSCI), 2017 IEEE Symposium Series on. pp. 1–6. IEEE (2017)
16. Loper, E., Bird, S.: Nltk: the natural language toolkit. arXiv preprint cs/0205028 (2002)
17. Martin, A.D., Quinn, K.M., Ruger, T.W., Kim, P.T.: Competing approaches to predicting supreme court decision making. Perspectives on Politics 2(4), 761–767 (2004)
18. Medvedeva, M., Vols, M., Wieling, M.: Judicial decisions of the european court of human rights: Looking into the crystal ball. In: Proceedings of the Conference on Empirical Legal Studies (2018)
19. Pedregosa, F., Varoquaux, G., Gramfort, A., Michel, V., Thirion, B., Grisel, O., Blondel, M., Prettenhofer, P., Weiss, R., Dubourg, V., Vanderplas, J., Passos, A., Cournapeau, D., Brucher, M., Perrot, M., Duchesnay, E.: Scikit-learn: Machine learning in Python. Journal of Machine Learning Research 12, 2825–2830 (2011)
20. Pennington, J., Socher, R., Manning, C.: Glove: Global vectors for word representation. In: Proceedings of the 2014 conference on empirical methods in natural language processing (EMNLP). pp. 1532–1543 (2014)
21. Řehůřek, R., Sojka, P.: Software Framework for Topic Modelling with Large Corpora. In: Proceedings of the LREC 2010 Workshop on New Challenges for NLP Frameworks. pp. 45–50. ELRA, Valletta, Malta (May 2010), http://is.muni.cz/publication/884893/en
22. Ruger, T.W., Kim, P.T., Martin, A.D., Quinn, K.M.: The supreme court forecasting project: Legal and political science approaches to predicting supreme court decisionmaking. Columbia Law Review pp. 1150–1210 (2004)
23. vizlegal: Features. https://www.vizlegal.com (May 2019)